\documentclass[letterpaper, 10 pt, conference]{ieeeconf}  

\IEEEoverridecommandlockouts                              

\usepackage{amsmath}
\usepackage{amssymb}
\usepackage{mathtools}

\usepackage{subcaption}
\usepackage{wrapfig}
\usepackage{graphicx}
\usepackage[font=small]{caption}

\usepackage{hyperref}

\usepackage{xcolor}

\usepackage{booktabs}

\newcommand{\name}{Goal-Contrastive Rewards}
\newcommand{\shortname}{GCR}

\title{\LARGE \bf
  On-Robot Reinforcement Learning with Goal-Contrastive Rewards
}

\author{Ondrej Biza\textsuperscript{1,2}, Thomas Weng\textsuperscript{1,*}, Lingfeng Sun\textsuperscript{1,*}, Karl Schmeckpeper\textsuperscript{1,*}, Tarik Kelestemur\textsuperscript{1,*}, \\ Yecheng Jason Ma$\textsuperscript{3,\textdagger}$, Robert Platt$\textsuperscript{1,2,\textdagger}$, Jan-Willem van de Meent$\textsuperscript{4,\textdagger}$, Lawson L. S. Wong$\textsuperscript{2,\textdagger}$%
\thanks{*Equal contribution: implementation and experiments. \textdagger Equal contribution: technical advising. \textsuperscript{1} Robotics and AI Institute, \textsuperscript{2} Northeastern University, Khoury College of Computer Sciences, \textsuperscript{3} University of Pennsylvania, \textsuperscript{4} University of Amsterdam. Correspondence to \texttt{biza.o@northeastern.edu}.}
}

\begin{document}

\maketitle
\thispagestyle{empty}
\pagestyle{empty}


\begin{abstract}
    Reinforcement Learning (RL) has the potential to enable robots to learn from their own actions in the real world. Unfortunately, RL can be prohibitively expensive, in terms of on-robot runtime, due to inefficient exploration when learning from a sparse reward signal. Designing dense reward functions is labour-intensive and requires domain expertise. In our work, we propose \name{} (\shortname{}), a dense reward function learning method that can be trained on passive video demonstrations. By using videos without actions, our method is easier to scale, as we can use arbitrary videos. \shortname{} combines two loss functions, an implicit value loss function that models how the reward increases when traversing a successful trajectory, and a goal-contrastive loss that discriminates between successful and failed trajectories. We perform experiments in simulated manipulation environments across RoboMimic and MimicGen tasks, as well as in the real world using a Franka arm and a Spot quadruped. We find that \shortname{} leads to a more-sample efficient RL, enabling model-free RL to solve about twice as many tasks as our baseline reward learning methods. We also demonstrate positive cross-embodiment transfer from videos of people and of other robots performing a task. Website: \url{https://gcr-robot.github.io/}.
\end{abstract}

\section{INTRODUCTION}

On-robot Reinforcement Learning (RL) of manipulation policies is highly challenging because of the exploration problem: a robot can spend hours randomly exploring its environment before chancing upon a goal state. %
Prior approaches to improve sample-efficiency for on-robot RL relied on %
hand-designing behavior priors~\cite{mendonca24continuously}, primitives~\cite{nasiriany22augmenting} and equivariant policies~\cite{zhu22sample}, or on bootstrapping RL with teleoperated demonstrations~\cite{ball23efficient,hu23imitation}.  Foundation models are effective at detecting goal completion~\cite{yang24robot,kumar24practice,brohan23rt2,du23vision}, but their application to fine-grained reward prediction is more challenging~\cite{huang23voxposer,yu23language}. Can we design a scalable way to guide RL exploration across many manipulation tasks?

The robotics community has had a growing interest in learning state similarities from passive videos (demonstrations without actions) \cite{sermanet18time,nair22r3m,ma23vip,ma23liv,ghosh23reinforcement,bhateja23robotic}.
These works learn an \textbf{implicit state value function} that takes a query image from a video and a goal image and estimates the similarity of their underlying states. 
Implicit state value functions can be trained using diverse and relatively abundant video data, including videos of a robot or human performing desired ``in-domain'' tasks, as well as videos of ``out-of-domain'' tasks performed by humans~\cite{grauman22ego4d,damen18scaling} and other robot morphologies~\cite{khazatsky24droid,openx}.
Therefore, implicit state value functions can scale much better than approaches requiring robot action data.
While implicit value pre-training has been primarily used as a representation learning method~\cite{nair22r3m,ma23vip,ma23liv,ghosh23reinforcement,bhateja23robotic}, its application to online RL is less explored~\cite{sermantet17unsupervised,sermanet18time}. 
In the context of RL, implicit state value functions could be used to scale reward learning by serving as an intrinsic reward for an RL agent; the agent is rewarded as its observed state progresses towards the desired goal state. %
Can implicit value functions effectively guide online RL?

\begin{figure}
    \centering
    \includegraphics[width=0.9\linewidth]{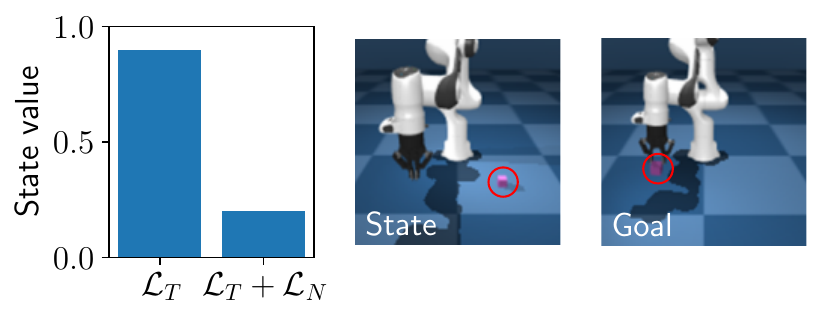}
    \caption{A model trained with only a temporal loss $\mathcal{L}_T$ assigns a high state value (left) to the pair of state-goal images (right) due to the arm being in the same pose, whereas a model trained with a combination of $\mathcal{L}_T$ and a contrastive loss $\mathcal{L}_N$ (ours) learns to distinguish the position of the block, assigning a low state value.}
    \label{fig:state_goal_values}
\end{figure}
\begin{figure}
    \centering
    \includegraphics[width=\linewidth]{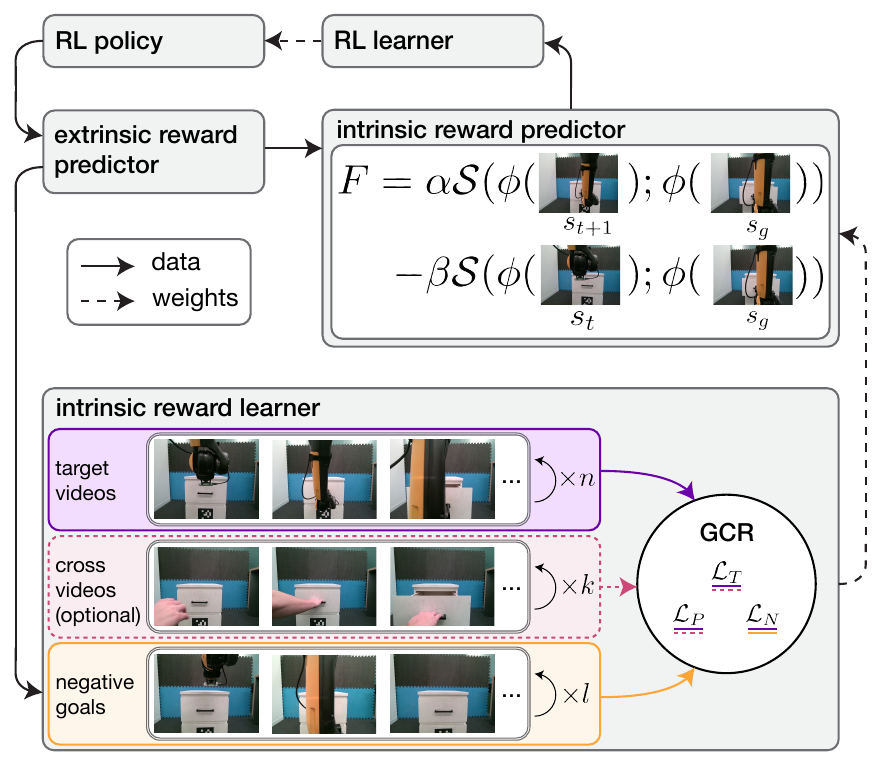}
    \caption{On-robot reinforcement learning system overview.}
    \vspace{-1em}
    \label{fig:system_diagram}
\end{figure}

In our work, we explore this question through the application of Value-Implicit Pre-training (VIP) \cite{ma23vip} to reward shaping for on-robot RL. The primary challenge we face is that throughout training the RL agent observes new states that might be out of distribution for the implicit value function. Consequently, it is possible (and often easy) for the agent to ``hack'' the value function by reaching high-value states that do not satisfy the goal condition. Figure \ref{fig:state_goal_values} shows one such example. We find that simply fine-tuning VIP on newly collected data is insufficient to learning an accurate value function.

To solve this problem, we propose a framework, \name{} (\shortname{}), that combines implicit value learning with a pair of contrastive losses over demonstrated and experienced goal states. The key intuition is that we should penalize recently visited states that achieve a high similarity to the goal (high state value) but do not satisfy the goal condition. Unlike \cite{fu18variational,singh19end}, which specifically focus on adversarial learning of goals, we leverage both our contrastive goal learning objective as well as the temporal-difference-like objective of VIP to propagate corrected state value estimates across time steps.

We perform experiments in simulated tasks from SERL \cite{luo24serl}, RoboMimic \cite{mandlekar21what} and MimicGen \cite{mandlekar23mimicgen}, a drawer opening with a real Boston Dynamics Spot, and picking and button pressing with a real tabletop Franka arm. Our comparison shows that \shortname{} enables efficient RL compared to learning with only sparse rewards and we further show a favorable comparison of \shortname{} to prior Inverse RL methods. \shortname{} enables real-world RL in around 1 to 2 hours, whereas learning with sparse rewards does not succeed. We also demonstrate that \shortname{} can be combined with other RL bootstrapping approaches and that we can benefit from cross-embodiment learning. Specifically, we show positive transfer between two different robot embodiments in simulation and between human and robot videos in the real world. On the systems side, our approach demonstrates an effective combination of intrinsic reward prediction and extrinsic reward prediction with foundation models as goal classifiers. We extend the asynchronous learning framework of \cite{luo24serl} to include implicit reward learning and intrinsic + extrinsic reward prediction that runs in parallel to RL control and learning.


In summary, our contributions are: 
\begin{enumerate}
    \item We propose a reward learning framework, \name{}, that learns a shaped reward function from passive videos and adapts to the recent behavior of an RL agent. \shortname{} significantly improves the sample-efficiency of on-robot RL both in sim and real experiments across two robots.
    \item We show that \shortname{} benefits from demonstrations of the tasks with different embodiments (robot and human).
    \item We demonstrate an asynchronous, on-robot RL system that combines intrinsic reward learning (\shortname{}) with extrinsic reward prediction (foundation models).
\end{enumerate}

\section{RELATED WORK}

\subsection{Inverse Reinforcement Learning}

Inverse Reinforcement Learning (IRL) aims to recover a reward function that explains a set of expert demonstrations. It primarily deals with the ambiguity of learning the right reward function given a limited coverage of the state space.  Early IRL works represent rewards as a linear combination of state features, $r(s; \theta) = \theta^T \phi(s)$, \cite{ziebart08maximum,abbeel04apprenticeship,ng00algorithms,russell98learning}. Later work has employed Gaussian Processes~\cite{levine11nonlinear} and neural networks~\cite{wulfmeier15maximum} to approximate the reward function. These approaches rely on an outer loop that learns $r(s; \theta)$ and an inner RL loop that learns a policy given the latest $\theta$, which might be too expensive for on-robot RL.

A second line of work frames IRL as an adversarial learning problem of distinguishing between state-action pairs $(s, a)$ or trajectories $(s_t, a_t)_{t=0}^T$ that either come from demonstrations or from the current policy~\cite{finn16guided,ho16generative,fu17learning,fu18variational,singh19end,reddy20sqil}. These works combine policy learning and reward function (or discriminator) learning in the same loop. Their general focus is on learning a reward function that matches the expert behavior in an unbiased way~\cite{fu17learning}; conversely, our focus is on learning a shaped reward that guides the agent towards goal states classified by an extrinsic reward function. We pursue learning from passive videos without actions due to their wide availability. Rank2Reward~\cite{yang24rank2reward} explores this setting by adapting prior Inverse RL methods to state-only trajectories. We use them as baselines.


\subsection{Learning from human videos}

A number of approaches attempt to translate videos of people into manipulation policies either directly or indirectly. Some methods detect the pose of the human hand and model the distribution of affordances and motion trajectories as a prior for robot policies~\cite{fang18demo2vec,xiong21learning,qin22dexmv,shaw22videodex,goyal22human,bahl22human,bharadhwaj23zeroshot,bahl23affordances,papagiannis24rx}. MimicPlay~\cite{wang23mimicplay} learns ``latent plans'' based on human hand trajectories as a conditioning for robot policies. Other approaches use in-painting to remove the human hand from videos~\cite{chang23look}, transform human videos into robot videos~\cite{smith20avid}, or align the latent representations of the videos~\cite{schmeckpeper2020reinforcement}. \cite{schmeckpeper2020learning,seo22reinforcement} trained a world model over both human and robot videos and \cite{patel22learning} reconstructed 3D models of objects found in people's hands in internet videos. We instead focus on using human videos to improve reward function learning for RL, making fewer assumptions about detecting and translating hand trajectories and detecting and understanding object interactions.

\section{PROBLEM FORMULATION}

We model an environment as a Markov Decision Process $M = \langle S, A, P, R, \gamma \rangle$ \cite{sutton18reinforcement}, where the reward function $R: S{\times}A{\times}S \rightarrow \mathbb{R}$ is sparse (zero for most states). We assume access to a dataset of $n$ passive (action-less) videos $P = \{(s^{(i)}_1, s^{(i)}_2, ...)\}_{i=1}^n$ and optionally $m$ demonstrations $D = \{ (s^{(i)}_1, a^{(i)}_1, r^{(i)}_1, s^{(i)}_2, a^{(i)}_2, r^{(i)}_2, ...) \}_{i=1}^m$ with actions and sparse rewards. We do not assume these datasets contain optimal or even successful behaviors, but we filter the data to only contain successful demonstrations in our experiments.

Our objective is to learn a policy $\pi_{\theta}$ that maximizes the expected return $\mathbb{E}_{\pi_{\theta}}\left[\sum_{t=0}^{\infty} \gamma^t r_t \right]$ while learning from a limited number of interactions with the environment. We use dataset $P$ to train a reward shaping network $F_{\phi}: S{\times}S \rightarrow \mathbb{R}$ that induces an MDP with a reward function $R'(s, a, s') = R(s, a, s') + F_{\phi}(s, s')$, where policy learning is faster.

\section{STATE SIMILARITY LEARNING}

State similarity (or implicit state value) learning methods use a dataset of videos $P = \{(s^{(i)}_1, s^{(i)}_2, ...)\}_{i=1}^n$ to learn a state similarity function $\mathcal{S}_{\phi}(s_1; s_2)$. The function can be implemented as a scaled dot product of encoded states, $\phi(s_1)^T\phi(s_2)$, \cite{ma23liv}, an $L_2$ distance between the encodings, $|| \phi(s_1) - \phi(s_2) ||_2$, \cite{ma23vip,sermanet18time} or as an additional neural prediction head \cite{yang24rank2reward}. Common loss functions for learning the state similarity function include time-contrastive learning \cite{sermanet18time}, learning to rank video frames \cite{yang24rank2reward} and action-less offline Reinforcement Learning \cite{ma23liv,ma23vip,ghosh23reinforcement}.

In our work, we use Value-implicit Pre-training (VIP) \cite{ma23vip}. VIP achieves action-less state value learning by optimizing a temporal-difference-like objective derived as a dual to a KL-regularized offline RL objective \cite{nachum19algaedice}:
\begin{align}
    &\mathcal{L}_{\textsc{VIP}} = \mathbb{E}_{p(g)} \left[ ( 1 \!-\! \gamma) \: \mathbb{E}_{\mu_0(o; g)}[-\mathcal{S}_{\phi}(o; g)] \right] \label{eq:vipi}\\ 
    &\quad+ \log \mathbb{E}_{(o, o'; g) \sim P} \left[ \exp \left\{ \mathcal{S}_{\phi}(o; g) - \delta(o) - \gamma \mathcal{S}_{\phi}(o'; g) \right\} \right].\nonumber
\end{align}
$p(g)$ is a distribution of goal images, $\mu_0$ is a distribution of initial states conditioned on a goal state and $\delta(o)$ is a goal indicator. In the absence of a reward function, goal states are sampled from the last few frames of each video. %

\section{GOAL-CONTRASTIVE REWARDS}


We propose \name{} (\shortname{}), a framework for learning dense reward functions for RL agents based on passive videos. The key contribution of our method is that it leverages both pre-training on arbitrary videos and online adaptation by fine-tuning as the RL agent reaches incorrect goals. First, we propose a combination of three losses that lead to a discriminative value function (Section \ref{sec:method:gcvf}). Second, we describe the conversion from a state value function to a dense reward function based on the theory of reward shaping (Section \ref{sec:method:rewards}). Third, we discuss the specifics of transferring values across embodiments, e.g. between two robots or between human and robot videos (Section \ref{sec:method:cross}). Fourth, we describe our extension of the SERL asynchronous RL library~\cite{luo24serl} that includes intrinsic and extrinsic reward functions (Section \ref{sec:method:async}).

\subsection{\name{} (\shortname{})}
\label{sec:method:gcvf}

We first make the observation that methods which learn to embed images based on their temporal distance in videos (e.g. R3M~\cite{nair22r3m}, VIP~\cite{ma23vip}, LIV~\cite{ma23liv} and ICVF~\cite{ghosh23reinforcement}) might not be suitable reward functions in online RL. These methods model the progress of successful behaviors towards the goal, but are not discriminative enough to penalize incorrect states that look similar to the goal state. For example, in Figure \ref{fig:state_goal_values}, we show that a model trained with a temporal loss $\mathcal{L}_T$ (VIP in this case) assigns a high value to a state where the robot matches the goal joint configuration without picking up a cube, thus failing the task. We note that RL agents tend to discover these gaps in the learned reward function, because they are easier to reach than the correct goal, leading to incorrect behaviors.


As a solution, we propose to use a contrastive loss consisting of a positive term $\mathcal{L}_P$ and a negative term $\mathcal{L}_N$ to learn the representation of goal images.
\begin{align}
    \mathcal{L} &= \mathcal{L}_T + \mathbb{E}_{(g, g', g_{\text{neg}})} \left[ \omega_1 \mathcal{L}_P(g, g') + \omega_2 \mathcal{L}_N(g, g_{\text{neg}}) \right] \label{eq:framework}
\end{align}
The positive term $\mathcal{L}_P$ pulls together the representation of similar goals $(g, g')$, whereas the negative term $\mathcal{L}_N$ decreases the similarity of a positive and negative goal image pair $(g, g_{\text{neg}})$. The effectiveness of this framework is highly dependent on the sampling of $g_{\text{neg}}$. In particular, the learning algorithm benefits from $g_{\text{neg}}$ that are close to the goal image in the representation space of $F_{\phi}$ but far from the goal in the ground-truth MDP (in other words, states that have a high predicted value but a low ground-truth value).

Crucially, we sample $g_{\text{neg}}$ from a small replay buffer that is updated with online data from the RL agent. Specifically, we sample the last $\beta$ frames in each failed episode (zero return) expressed as the fraction of the length of the episode. Our learned reward function always adapts to the recent, incorrect, behaviors of the RL agent. This leads to a positive feedback loop, where the RL agent discovers gaps in the learned reward function and the collected data are immediately used to improve the reward model.


In our experiments, we instantiate Equation $\ref{eq:framework}$ with the VIP loss as $L_T$ and with a simple contrastive (SC) loss as $L_P$ and $L_N$. In the following equation, $S_{\phi}$ is the cosine similarity of images encoded by $\phi$.
\begin{align}
    \begin{split}
    &\mathcal{L}_{\textsc{\shortname{}(SC)}} = \\
    &\quad\mathcal{L}_{\textsc{VIP}} + \mathbb{E}_{(g, g', g_{\text{neg}})} \left[ - \omega_1 \mathcal{S}_{\phi}(g; g') + \omega_2 \mathcal{S}_{\phi}(g; g_{\text{neg}}) \right].
    \end{split}
\end{align}
We also experiment with an InfoNCE-like contrastive (IC) loss \cite{oord18representation}. $\mathcal{L}_{\textsc{\shortname{}(SC)}}$ is an upper bound on  $\mathcal{L}_{\textsc{\shortname{}(IC)}}$.
\begin{align}
    \begin{split}
    &\mathcal{L}_{\textsc{\shortname{}(IC)}} = \\
    &\quad
    \mathcal{L}_{\textsc{VIP}} + \mathbb{E}_{(g, g')} \left[ - \log \frac{\exp \{ \omega_1 \mathcal{S}_{\phi}(g; g') \} }{ \mathbb{E}_{g_{\text{neg}}} \left[ \exp \{ \omega_2 \mathcal{S}_{\phi}(g; g_\text{neg}) \} \right] } \right].
    \end{split}
\end{align}

In summary, the proposed loss function $\mathcal{L}$ trains a state similarity function $\mathcal{S}_{\phi}(s_1, s_2)$ based on passive videos $P$ as well as replay buffer data from online RL. Next, we describe turning $\mathcal{S}_{\phi}(s_1, s_2)$ into a reward function.

\subsection{Dense reward functions}
\label{sec:method:rewards}

We formulate our reward function as the sum of the sparse reward $R$ and a reward shaping term $F$. Based on the theoretical analysis of Ng, Harada and Russel~\cite{ng99policy}, $F$ is formulated as the difference of the potential $\Phi$ of the next state $s'$ and the current state $s$.
\begin{align}
    \begin{split}
    R'(s, a, s') &= R(s, a, s') + F(s, s') \\
    &= R(s, a, s') + \alpha \Phi(s';g) - \beta \Phi(s;g)
    \end{split}
\end{align}

Depending on the setting of $\alpha$ and $\beta$, we find various reward shaping terms from prior works:
\begin{enumerate}
    \item $\alpha=1, \beta=1$: a simple difference between the potential of the current and next state \cite{ma23vip,li22phasic,lee21generalizable}. It is a version of the potential-based reward function from \cite{ng99policy} with a slight bias of $(1 - \gamma) \Phi(s';g)$. \\
    \item $\alpha=1, \beta=0$: prior inverse RL works commonly use only the potential of the next state as a reward function \cite{yang24rank2reward,fu18variational,ho16generative}; this reward might be susceptible to the agent getting stuck in endless loops \cite{ng99policy,randlov98learning}. \\
\end{enumerate}

\noindent We use the first option in our work. We implement the potential $\Phi$ as a re-scaled cosine similarity $\mathcal{S}_{\phi}$ of images encoded with $\phi$, learned with the loss functions described in Section \ref{sec:method:gcvf}. This potential can be interpreted as the discount factor $\gamma$ to the power of the number of steps to go before reaching a goal under some behavioral policy $\pi_b$:
\begin{align}
    \Phi(s; g) = (\mathcal{S}_{\phi}(s; g) + 1) / 2
    \approx \mathbb{E}_{\pi_b} \left[ \sum_{t=0}^{\infty} \gamma^t \text{is\_goal}(s_t) \right]
\end{align}

\begin{figure}[t]
    \centering
    \vspace{0.6em}
    \includegraphics[width=\linewidth]{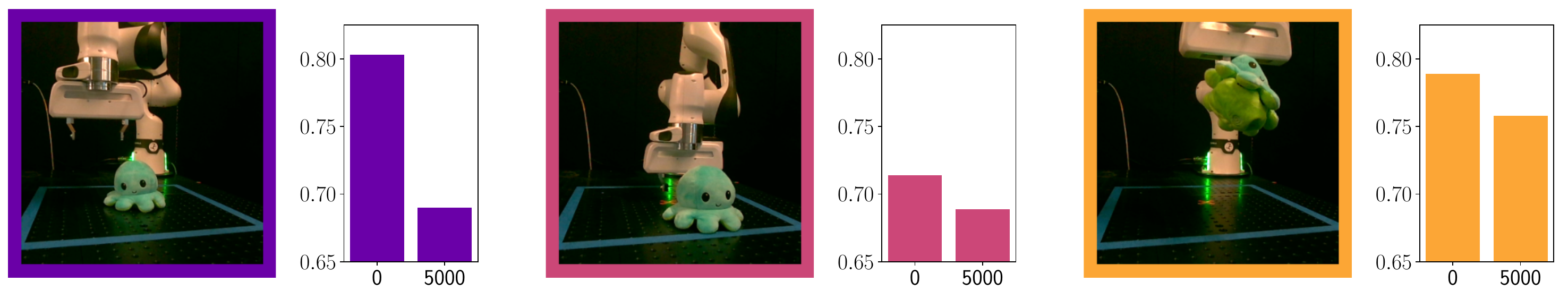}
    \vspace{-0.6em}
    \caption{We show three example states and their associated GCR state values (scaled zero to one) at 0 and 5000 online \shortname{} training steps. The 0 step version of \shortname{} is fine-tuned on demonstrations, but not on any online data. 5000 steps correspond to approximately one hour of on-robot training.}
    \label{fig:lift_values}
    \vspace{-1em}
\end{figure}

\subsection{Cross-embodiment learning}
\label{sec:method:cross}

We have $n$ passive video demonstrations with the target embodiment and further $k$ in-domain videos with a different embodiment (e.g. different robotic arm or videos of people). Usually, $k$ is many times larger than $n$ (e.g. it is much easier to collect videos of people). First, we oversample the target embodiment videos so that they have the same parity as the other embodiment videos. Second, we find that the positive loss $\mathcal{L}_P$ is sufficient for positive transfer across embodiments, as it pulls together the encoded images of goal states across episodes invariant to the appearance of the manipulator. During inference, we only use goal states from the target embodiment videos. Third, we also benefit from a ``negative'' dataset sampling step in VIP (Appendix D1 in \cite{ma23vip}) that applies TD backups to transitions with goal images randomly sampled from the dataset. This leads to backups that combine images from both embodiments.

\subsection{Parallel reward and RL training}
\label{sec:method:async}


We base our system implementation on the SERL library~\cite{luo24serl}, which implements the RL loop as two parallel processes that perform action execution and RL. ZeroMQ~\cite{zeromq} is used for performant messaging between processes. Our system implements five processes that run in parallel: RL actor, RL learner, intrinsic reward learner, intrinsic reward predictor and extrinsic reward predictor (Figure \ref{fig:system_diagram}). The RL actor and learner come from the original SERL implementation. The intrinsic reward learner runs \shortname{} learning and the intrinsic reward predictor evaluates \shortname{} rewards using the latest checkpoint from the learner. The extrinsic reward predictor runs various foundation models to evaluate if a goal condition was reached; it is only used in real-world experiments. This distributed system provides a significant benefit in terms of real-world experiment runtime, as we can commit extra computation to reward learning and prediction without limiting the speed of policy execution.

\begin{figure}[t]
    \centering
    \begin{subfigure}[b]{0.24\linewidth}
        \centering
        \includegraphics[width=\textwidth]{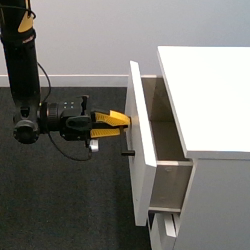}
    \end{subfigure}
    \begin{subfigure}[b]{0.24\linewidth}
        \centering
        \includegraphics[width=\textwidth]{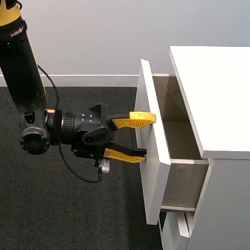}
    \end{subfigure}
    \begin{subfigure}[b]{0.24\linewidth}
        \centering
        \includegraphics[width=\textwidth]{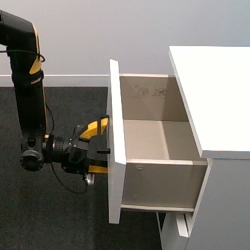}
    \end{subfigure}
    \begin{subfigure}[b]{0.24\linewidth}
        \centering
        \includegraphics[width=\textwidth]{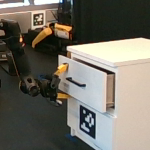}
    \end{subfigure}
    \caption{Four different drawer opening policies learned by \shortname{}. First: handle grasp, second: top grasp, third: top finger through handle, fourth: bottom grasp. We find that \shortname{} improves exploration but does not prescribe a specific way of opening the drawer, whereas RLPD always converges to the same policy (handle grasp).}
    \label{fig:rw_demo}
\end{figure}

\section{EXPERIMENTS}

\begin{figure*}
    \centering
    \begin{subfigure}[b]{0.65\textwidth}
        \centering
        \includegraphics[width=\textwidth]{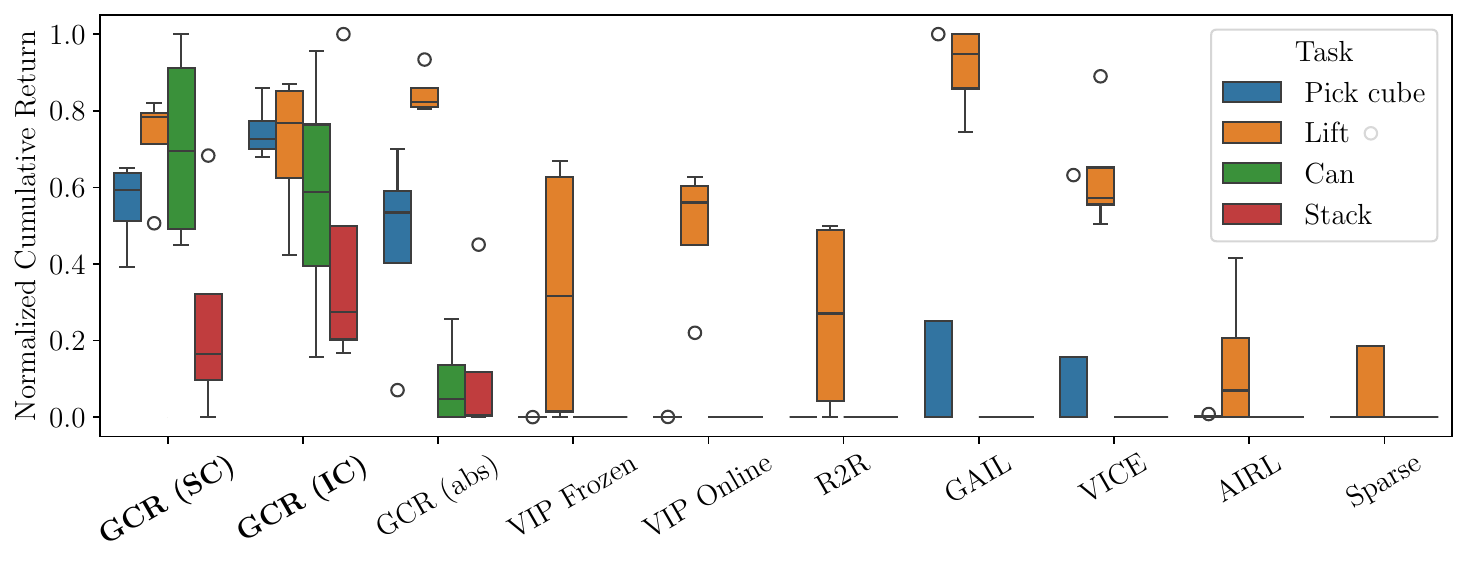}
    \end{subfigure}
    \begin{subfigure}[b]{0.11\textwidth}
        \centering
        \includegraphics[width=1.07\textwidth]{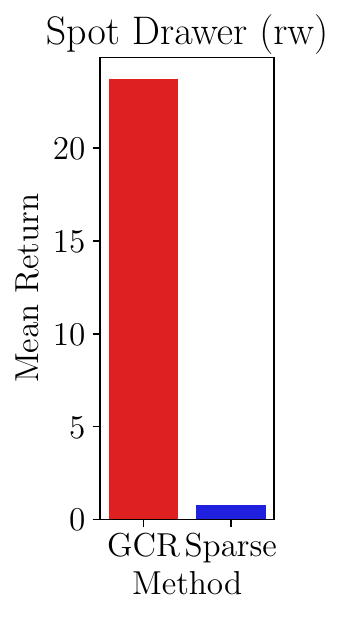}
        \vspace{0.4em}
    \end{subfigure}
    \begin{subfigure}[b]{0.11\textwidth}
        \centering
        \includegraphics[width=1\textwidth]{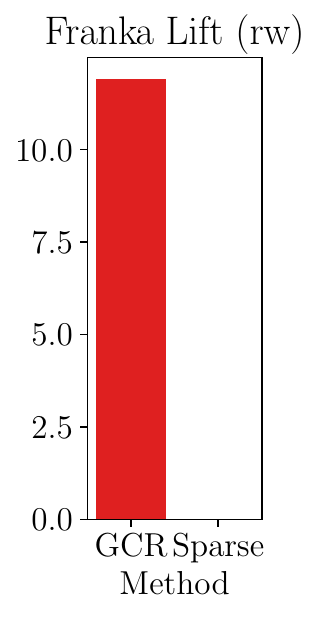}
        \vspace{0.4em}
    \end{subfigure}
    \begin{subfigure}[b]{0.11\textwidth}
        \centering
        \includegraphics[width=1.04\textwidth]{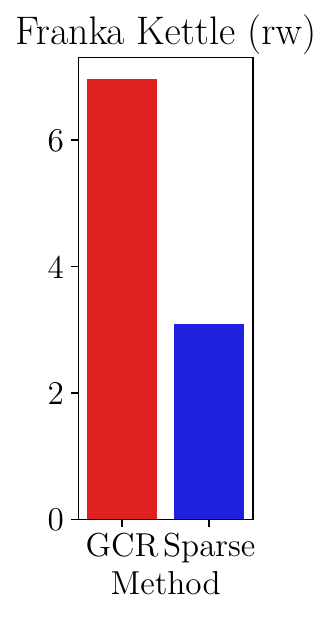}
        \vspace{0.4em}
    \end{subfigure}
    \caption{(Left) Cumulative returns of DrQ trained with different reward functions (x axis). We normalize the cumulative returns by the highest achieved value across all methods and runs. We report four random seeds across three simulated tasks, SERL Pick cube (20 \textit{passive} demos), RoboMimic Lift (20) and Can (20), and MimicGen Stack D0 (100). (Right) We also add results for real-world Franka Lift, Kettle and Spot drawer opening tasks (Figure \ref{fig:rw_tasks}). We run these tasks with only \shortname{} and Sparse rewards.}
    \label{fig:model_free_comparison}
    \vspace{-1em}
\end{figure*}

\begin{figure}
    \centering
    \begin{subfigure}[b]{0.4\linewidth}
        \centering
        \includegraphics[width=\textwidth]{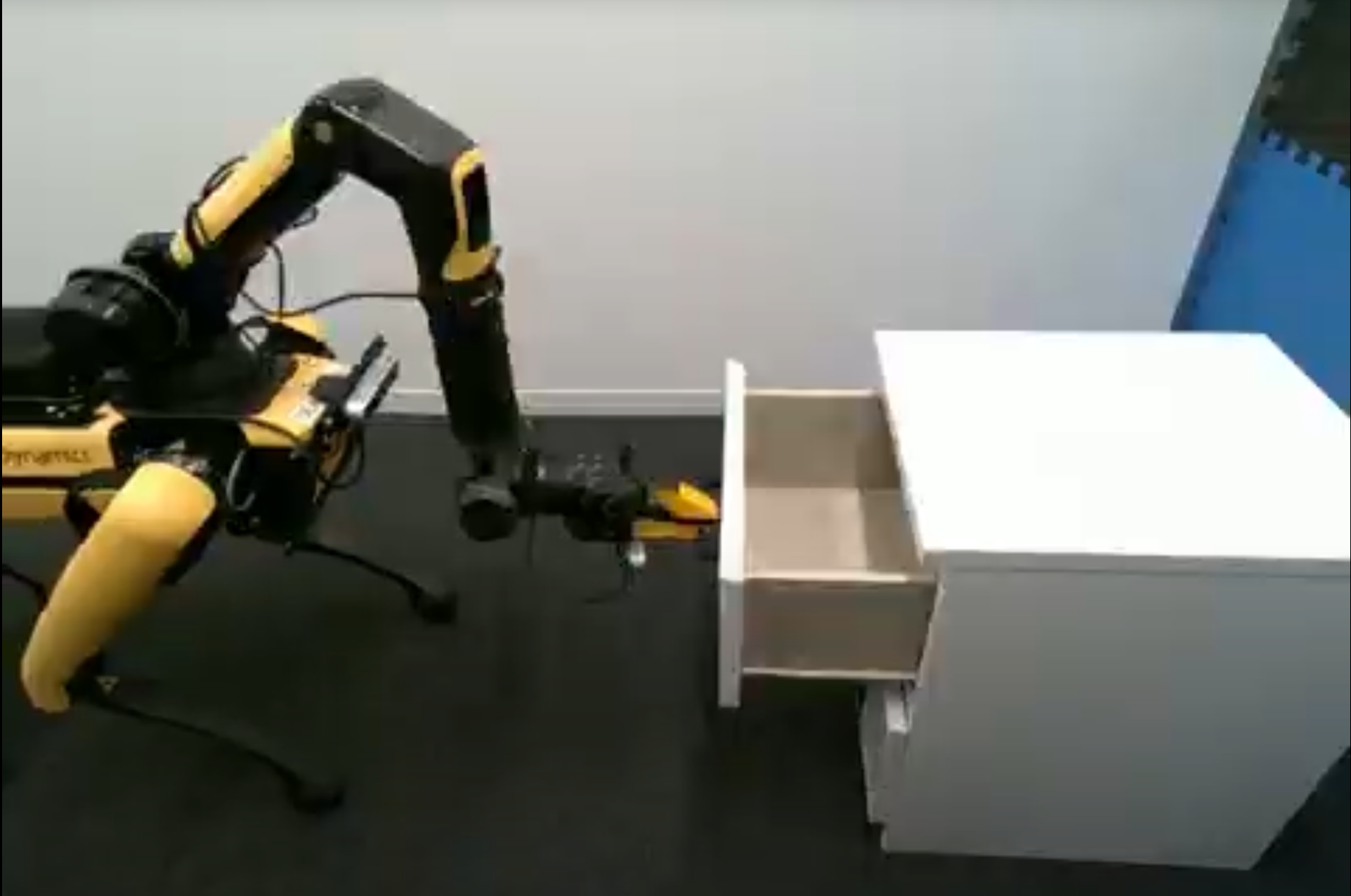}
    \end{subfigure}
    \centering
    \begin{subfigure}[b]{0.275\linewidth}
        \centering
        \includegraphics[width=\textwidth]{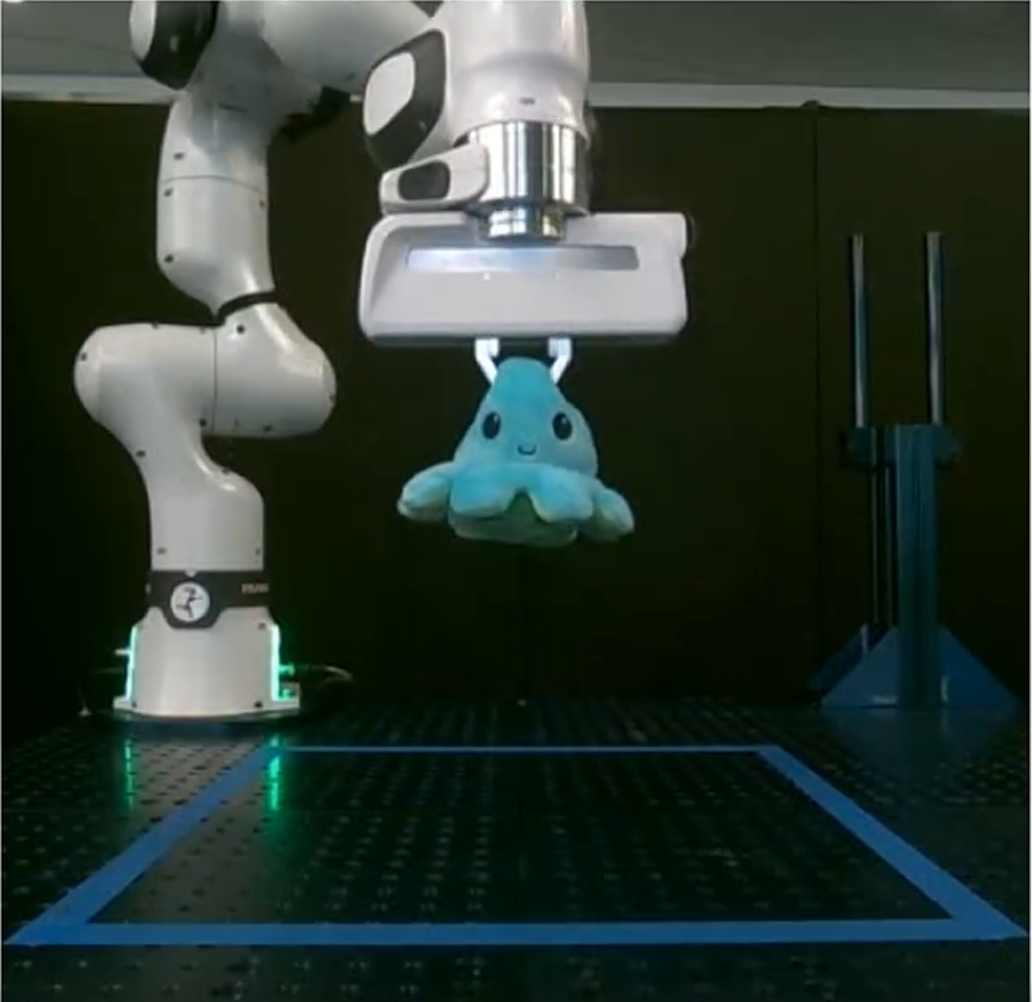}
    \end{subfigure}
    \centering
    \begin{subfigure}[b]{0.267\linewidth}
        \centering
        \includegraphics[width=\textwidth]{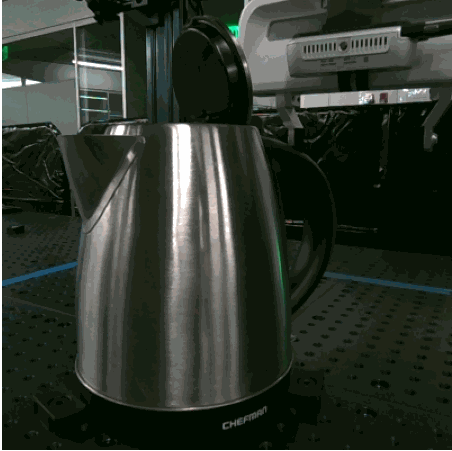}
    \end{subfigure}
    \caption{Real-world reinforcement learning task: Spot drawer opening, Franka plushie picking and Franka kettle opening.}
    \label{fig:rw_tasks}
    \vspace{-1em}
\end{figure}


We perform online Reinforcement Learning (RL) experiments with SERL, RoboMimic and MimicGen simulated tasks and two real-world robot platforms to answer the following questions:

\begin{enumerate}
    \item Does \shortname{} lead to efficient online RL in domains with sparse rewards (Section \ref{exp:model_free})?
    \item Can we combine \shortname{} with reinforcement learning from demonstrations (Section \ref{exp:rlpd})?
    \item Does \shortname{} benefit from videos of other embodiments (Section \ref{exp:cross})?
\end{enumerate}


\textbf{Simulation:} We adapt SERL Pick cube \cite{luo24serl}, RoboMimic Lift, Can and Square \cite{mandlekar21what} and MimicGen Stack D0 and Coffee D0 \cite{mandlekar23mimicgen} to a reinforcement learning setting with sparse rewards. The observation space consists of two RGB images (shoulder and wrist camera) and robot proprioception. The action space represents the delta movements of the end-effector and a gripper open/close command. All environments use a Franka arm; we also collect cross-embodiment data (Section \ref{sec:method:cross}) with a Kuka arm.

\textbf{Physical robots:} We perform real-world experiments with a table-top Franka and a Boston Dynamics Spot with a back-mounted arm and a Fin Ray style gripper. The Franka tasks are picking up a plushie and opening a kettle by pushing a button. The Spot task is to open a drawer. We use a pair of front and wrist cameras in all experiments. We use impedance and admittance control on the Franka and the Spot respectively to limit the forces on the end-effector. The extrinsic rewards (which predict if we have reached a goal state) are implemented with foundation models: we use Grounding DINO \cite{liu23grounding} with Segment Anything \cite{kirillov23segment} to detect and mask the plushie and the drawer. We predict success heuristically based on the mask position and size. For kettle opening, we query GPT-4 vision \cite{openai23gpt4} with a yes/no question.

\textbf{Baselines and ablations:} We compare our method to previous Inverse Reinforcement Learning methods. Specifically, Rank2Reward (R2R) \cite{yang24rank2reward} and a version of Generative Adversarial Imitation Learning (GAIL) \cite{ho16generative}, Variational Inverse Control with Events (VICE) \cite{fu17learning} and Adversarial Inverse Reinforcement Learning (AIRL) \cite{fu17learning} adapted to passive videos. We further compare \shortname{} to only using sparse rewards (Sparse) and to using sparse rewards with RLPD \cite{ball23efficient} (Sparse + RLPD). RLPD requires demonstrations with actions. We consider five ablations of our method. VIP Frozen: a reward function derived from VIP fine-tuned on passive demonstrations and frozen during RL training; VIP Online: VIP fine-tuned on both passive demonstrations and online data from the RL agent; \shortname{} (abs) uses the absolute reward (Section \ref{sec:method:rewards}, $\alpha = 0, \beta = 1$) instead of a delta reward; \shortname{} (IC) uses an InfoNCE contrastive loss instead of the Simple Contrastive (\shortname{} (SC)) loss. We use an unfrozen ResNet-50 LIV backbone \cite{ma23liv} pre-trained on EPIC-KITCHENS \cite{damen18scaling} in all our experiments unless stated otherwise. Further details are in the Appendix \ref{appendix}.

\newcommand{\wcross}{0.24}
\begin{figure*}
    \centering
    \centering
    \begin{subfigure}[b]{\wcross \textwidth}
        \centering
        \includegraphics[width=\textwidth]{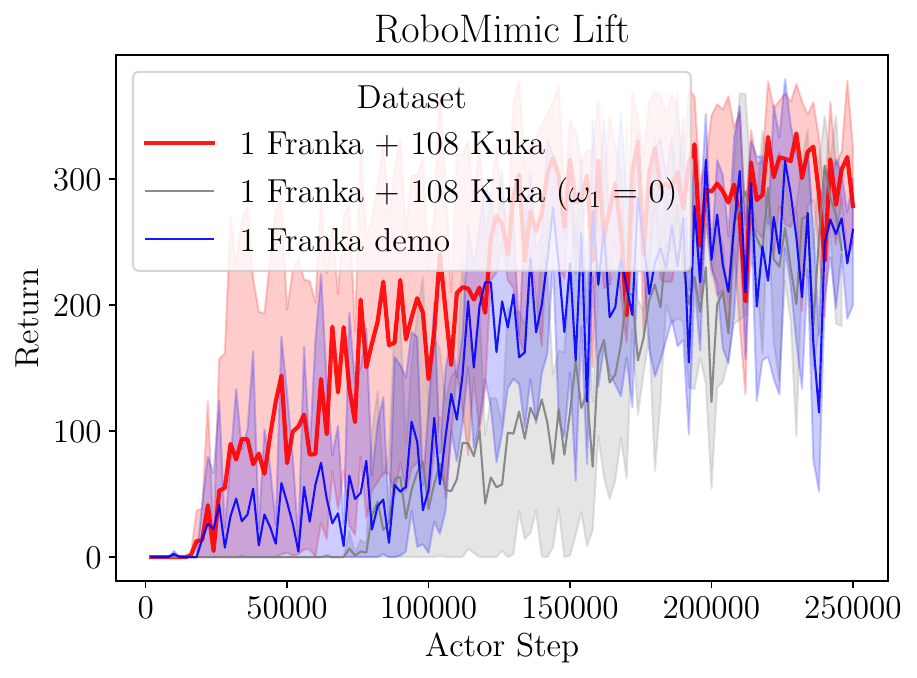}
    \end{subfigure}
    \begin{subfigure}[b]{\wcross \textwidth}
        \centering
        \includegraphics[width=\textwidth]{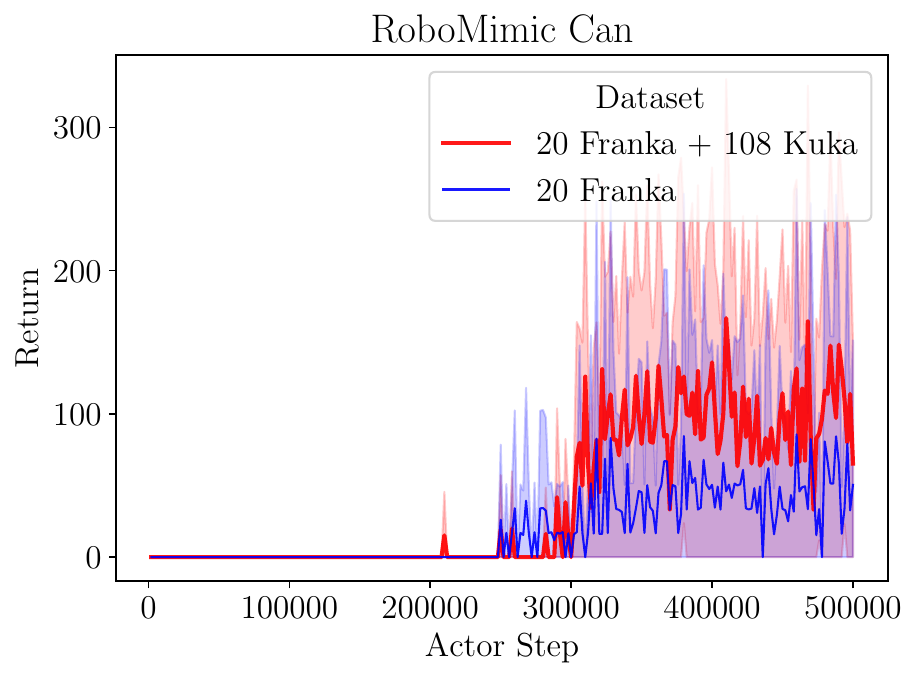}
    \end{subfigure}
    \begin{subfigure}[b]{\wcross \textwidth}
        \centering
        \includegraphics[width=\textwidth]{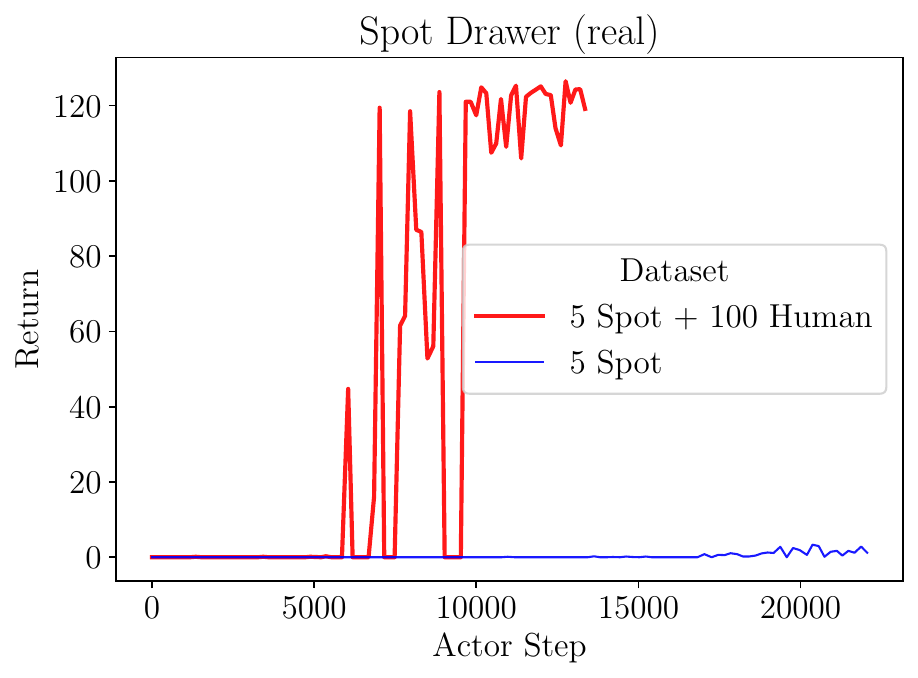}
    \end{subfigure}
    \begin{subfigure}[b]{\wcross \textwidth}
        \centering
        \includegraphics[width=\textwidth]{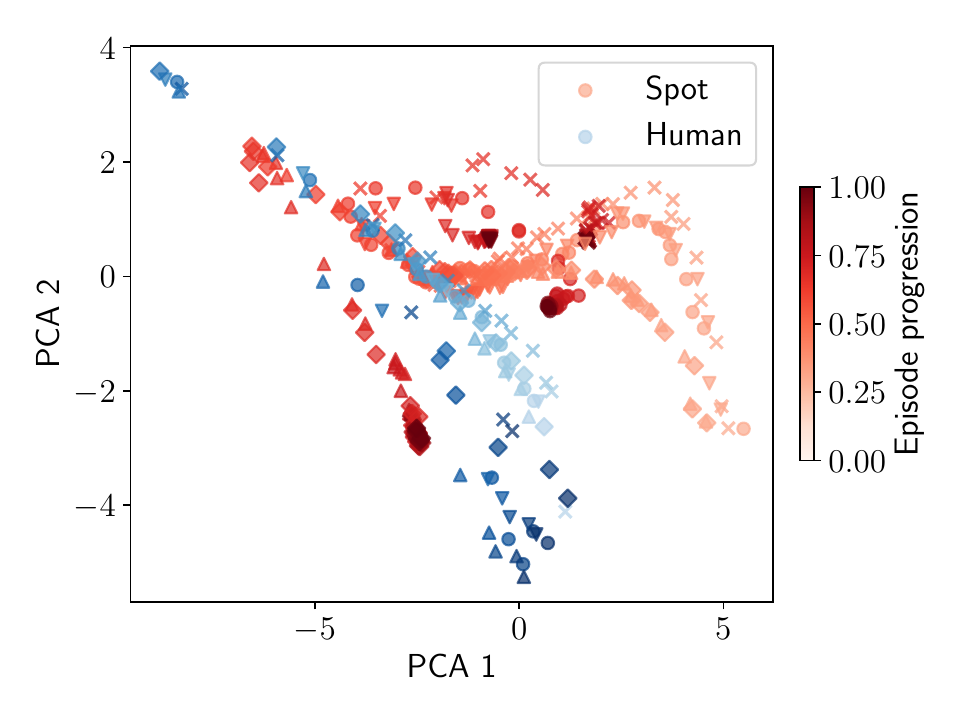}
    \end{subfigure}
    \caption{Episode returns for DrQ trained with \shortname{} reward functions. \shortname{} is provided with either only target domain demonstrations (blue line) or with a combination of target and cross-embodiment demonstrations (red line). The fourth panel show a latent space of a \shortname{} model with encoded images from human (blue) and robot (red) demonstrations. Each symbol marks a different episode and the brightness of the point represent the episode time step.}
    \label{fig:cross_comparison}
    \vspace{-1em}
\end{figure*}

\subsection{Model-free RL with learned reward shaping}
\label{exp:model_free}


We first measure the impact of reward shaping using \shortname{}, our baselines, and ablations on model-free reinforcement learning. Figure \ref{fig:model_free_comparison} shows that \shortname{} learns all four tasks (SERL Pick cube, RoboMimic Lift and Can, MimicGen Stack D0) with 4/4 seeds eventually converging in Pick cube, Lift and Can, and 1/4 seeds converging in Stack. In contrast, R2R, GAIL, VICE, AIRL and Sparse rewards generally only lead to success in Lift. While Rank2Reward (R2R) performs well on MetaWorld \cite{yu19metaworld} tasks in \cite{yang24rank2reward}, we did not see the same benefit in RoboMimic and MimicGen tasks. To investigate this further, we ran R2R with four backbones -- R3M \cite{nair22r3m} frozen/unfrozen and LIV \cite{ma23liv} frozen/unfrozen -- and we report the best result. VIP Frozen and VIP Online, which use only the $\mathcal{L}_T$ component of our loss function, only learn Lift, underscoring the importance of goal-contrastive learning. Pick cube and Lift both involve picking up a cube, with a different camera angle and environment appearance. Since the cube looks smaller in Pick cube, we hypothesize that the baselines tend to ignore the cube in favour of overfitting on the robot movements.

The InfoNCE variant of our method (\shortname{} (IC)) performs similarly to the Simple Contrastive variant (\shortname{} (SC)). Using absolute rewards (\shortname{} (abs)) leads to comparable performance on the simple tasks (Pick cube and Lift), but delta rewards (used in \shortname{} (SC) and (IC)) significantly outperform on the difficult tasks (Can and Stack). Finally, we show a comparison between Sparse and \shortname{} rewards in three real-world experiments in Figure \ref{fig:model_free_comparison}, right. We find that \shortname{} successfully converges in one to two hours in all three tasks: drawer opening, lifting a plushie, and opening a kettle.

\subsection{Bootstrapped RL}
\label{exp:rlpd}

In the previous section, we focused on using action-less videos to learn a shaped reward function. But, there are other approaches to bootstrapping RL exploration. We chose RLPD, a simple method that combines model-free RL and demonstrations with actions. Here, we use the same videos with actions for RLPD and without actions for GCR ($n=m$). Future work could explore a setting where we have passive videos with a few videos with actions ($n>m$). We examine the following question: can \shortname{} reward function learning synergize with RLPD replay buffer bootstrapping?

In Table \ref{tab:rlpd}, we list the mean returns of Sparse rewards + RLPD or \shortname{} + RLPD on the following tasks: SERL Pick cube, RoboMimic Lift, Can and Square and MimicGen Stack D0 and Coffee D0. Note that we added RoboMimic Square and MimicGen Coffee D0, which were too difficult to solve while learning from action-less demonstrations only.

\begin{table}[]
    \centering
    \begin{tabular}{cccc}
        \toprule
        Task & Num. Demos & Sparse + RLPD & \textbf{GCR + RLPD} \\
        \midrule
        Pick cube & 1 & 0.0 $\pm$ 0.0 & \textbf{28.3 $\pm$ 7.3} \\
        Pick cube & 5 & \textbf{54.6 $\pm$ 6.2} & \textbf{53.4 $\pm$ 2.1} \\
        Lift & 1 & 28.2 $\pm$ 42.0 & \textbf{306.8 $\pm$ 11.1} \\
        Lift & 5 & \textbf{271.7 $\pm$ 16.1} & \textbf{287.1 $\pm$ 21.7} \\
        Can & 20 & 0.0 $\pm$ 0.0 & \textbf{89.3 $\pm$ 28.9} \\
        Can & 100 & 0.0 $\pm$ 0.0 & \textbf{91.2 $\pm$ 22.9} \\
        Square & 110 & 0.0 $\pm$ 0.0 & \textbf{22.6 $\pm$ 27.0} \\
        Stack D0 & 20 & 0.0 $\pm$ 0.0 & \textbf{107.6 $\pm$ 21.7} \\
        Stack D0 & 100 & 0.0 $\pm$ 0.0 & \textbf{100.1 $\pm$ 28.9} \\
        Coffee D0 & 100 & \textbf{0.6 $\pm$ 0.8} & \textbf{7.7 $\pm$ 8.4} \\
        \bottomrule
    \end{tabular}
    \caption{Mean returns of GCR (ours) + RLPD compared to Sparse rewards + RLPD. Mean and standard deviations over four random seeds are reported. Bold numbers are within one standard deviation of the best result. Figure \ref{fig:rldp_example} shows an example learning curve.}
    \label{tab:rlpd}
    \vspace{-1em}
\end{table}

Our results show that \shortname{} greatly increases the sample-efficiency of RLPD, allowing it to learn Pick cube and Lift with a single demonstration (used by both RLPD and \shortname{}). RLPD cannot solve RoboMimic Can, Square, Stack and Coffee with 100+ demonstrations due to the long-horizon nature of the tasks, but \shortname{} enables RLPD to solve Can in 4/4 seeds, Stack in 4/4 and Square in 2/4. \shortname{} occasionally reaches the goal of Coffee D0 but does not converge in 500k environment steps.

We further visualize an interesting qualitative difference between \shortname{} and RLPD policies in Figure \ref{fig:rw_demo}. \shortname{} + DrQ (without RLPD) learns four different ways of opening the drawer across separate runs. Conversely, RLPD always converges to the demonstrated policy, which involves opening the drawer by grasping its handle. We believe that reward shaping (with \shortname{}) is less descriptive in terms of what policy should be learned, compared to RLPD, which directly feeds demonstrations with actions into the model-free RL replay buffer. This can be a drawback if we want the RL agent to closely follow the demonstration, but also a benefit if we want to learn a more diverse set of policies or possibly find a better way of performing a task.

\subsection{Cross-embodiment learning}
\label{exp:cross}

An additional benefit of \shortname{} is the ability to learn from videos from other embodiments. We collect 108 videos of a Kuka robot in RoboMimic Lift and Can, and 100 videos of real-world drawer opening with three human participants. We find that cross-embodiment learning benefits all three tasks (Figure \ref{fig:cross_comparison}). The difference is more pronounced in the drawer opening experiment. We suspect that the real-world data contains more noise due to lighting changes, object movements, camera pose changes, etc., which make it so that \shortname{} greatly benefits from a larger dataset of human drawer opening. We also ablate \shortname{} by setting the positive loss weight to zero ($\omega_1 = 0)$, which negates the benefit of cross-embodiment data (Figure \ref{fig:cross_comparison}, first panel). Finally, we visualize the \shortname{} latent space of demonstrations across human and Spot episodes in Figure \ref{fig:cross_comparison}, fourth panel. We find that the embeddings especially overlap in the middle of the episode, when the drawer is being pulled open.

\begin{figure}
    \centering
    \begin{subfigure}[b]{0.48 \linewidth}
        \centering
        \includegraphics[width=\textwidth]{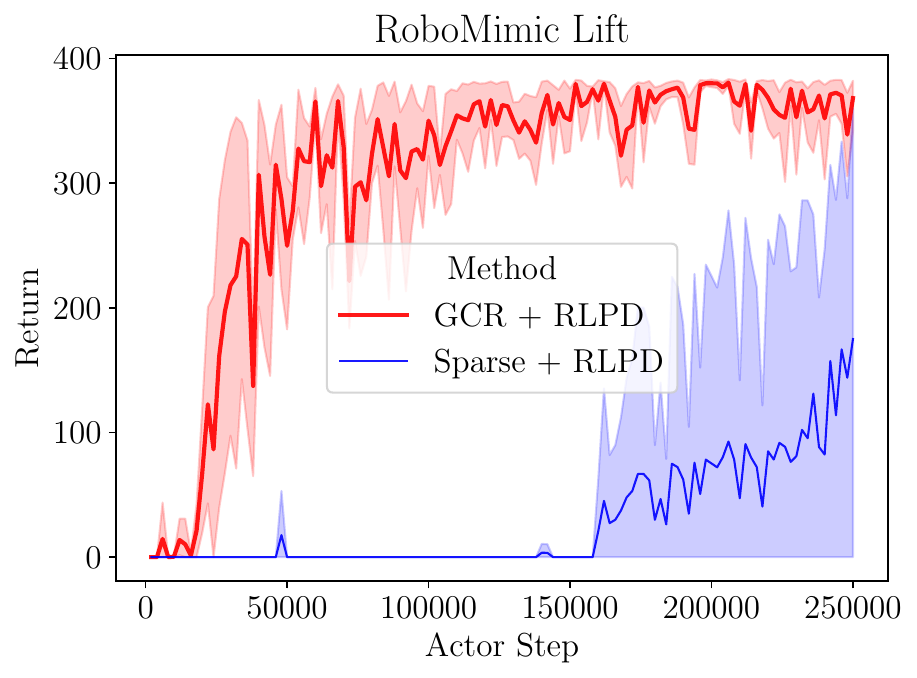}
    \end{subfigure}
        \begin{subfigure}[b]{0.48 \linewidth}
        \centering
        \includegraphics[width=0.3\textwidth]{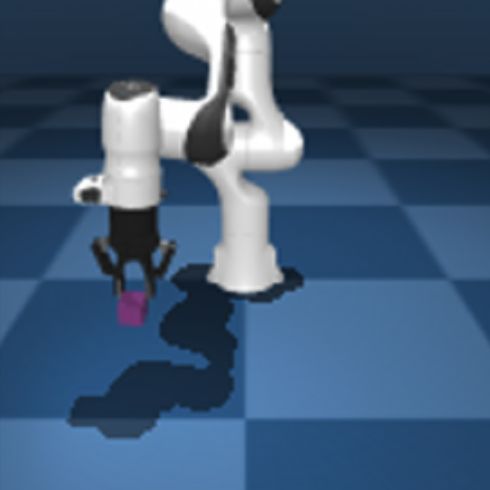}
        \includegraphics[width=0.3\textwidth]{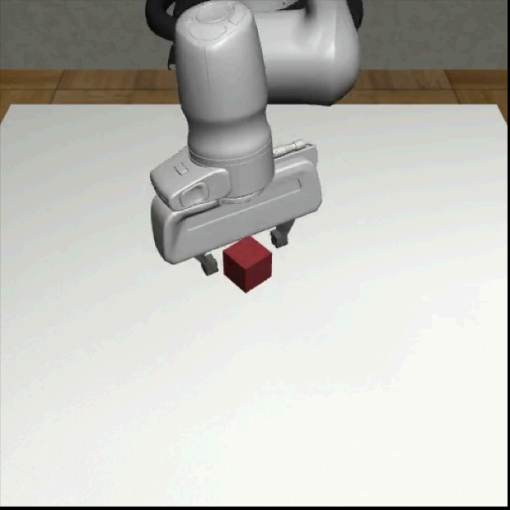}
        \includegraphics[width=0.3\textwidth]{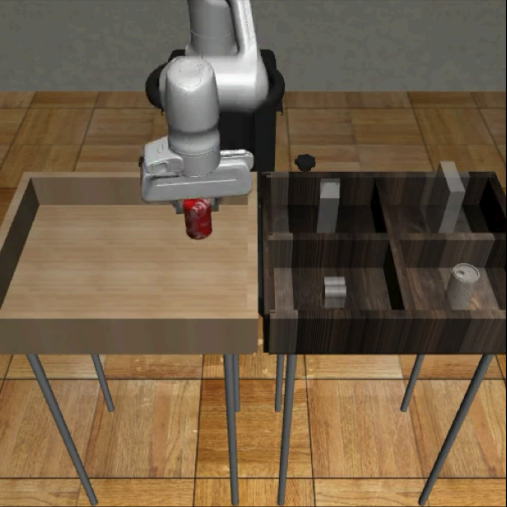}
        \vspace{1em}
        \includegraphics[width=0.3\textwidth]{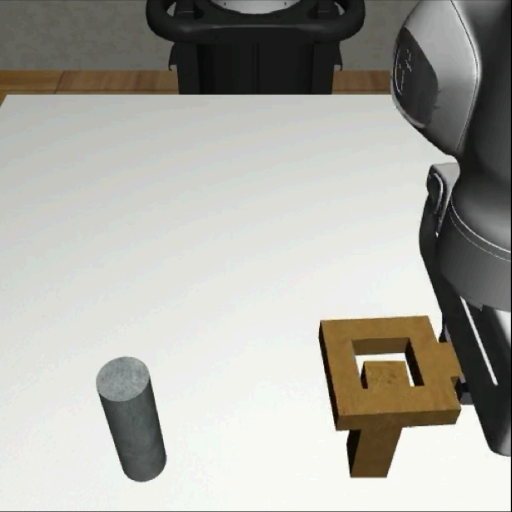}
        \includegraphics[width=0.3\textwidth]{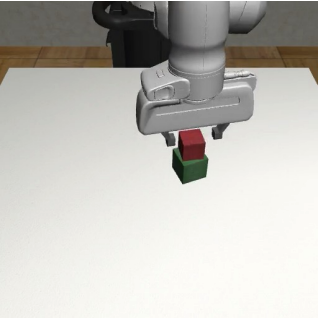}
        \includegraphics[width=0.3\textwidth]{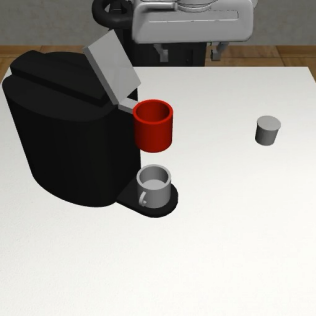}
    \end{subfigure}
    \caption{(Left) Example learning curve of RLPD with learned rewards (GCP) and sparse rewards (Sparse). RoboMimic Lift with one demonstration. Table \ref{tab:rlpd} lists quantitative results for RLPD. (Right) Simulated tasks: SERL Pick cube, RoboMimic Lift, Can and Square, MimicGen Stack D0 and Coffee D0.}
    \label{fig:rldp_example}
    \vspace{0.35em}
\end{figure}

\section{CONCLUSION}

\name{} provide an effective way of learning a shaped reward from passive videos. The learned dense reward function can then guide  the Reinforcement Learning of visuomotor policies in both simulation and the real world, greatly increasing its sample efficiency. We demonstrated this effect across six simulated tasks and two real robots, a Franka arm and a Spot quadruped, with a comparison to prior Inverse Reinforcement Learning methods. We also showed positive cross-embodiment transfer with videos from different embodiments. Altogether, we believe \shortname{} opens an interesting avenue of scaling up on-robot RL to many tasks. 

\textbf{Limitations and future work:} Our experiments are limited to learning a reward function from a single camera that is fixed in all experiments except for the Spot drawer opening. It is crucial to develop approaches that can effectively predict rewards from arbitrary views; we note prior works on view-invariant representation learning: \cite{sermanet18time,xue23learning,huang24egoexolearn}. For our RL agent, we use implementations of DrQ \cite{yarats21image} and RLPD \cite{ball23efficient} from prior work \cite{luo24serl} that have small convolutional networks as backbones, which might struggle to generalize across objects and environments. Future works could leverage improvements in RL architectures and hyper-parameter tuning \cite{hu23imitation} and in the pre-training of visual backbones \cite{majumdar23where,oquab24dinov2,shang24theia}. Finally, \shortname{} could be directly applied to multi-goal or multi-task learning.

\clearpage
\section*{ACKNOWLEDGMENT}

This work is supported in part by NSF 1750649, NSF 2107256, NSF 2314182, NSF 2409351 and NASA 80NSSC19K1474. Yecheng Jason Ma is supported by the Apple Scholars in AI/ML PhD Fellowship. We would like to thank Osman Dogan Yirmibesoglu for designing and manufacturing a flexible Fin-ray style griper for the Spot that we use in our drawer opening experiment. We would also like to thank Andy Park for creating the teleoperation system that we use for data collection and Kuan Fang for helpful discussion.





\bibliographystyle{IEEEtran}
\bibliography{main}


\clearpage
\section*{APPENDIX}
\label{appendix}

\subsection{Simulated Environment Details}

We adapt SERL Pick cube \cite{luo24serl}, RoboMimic Lift, Can, and Square \cite{mandlekar21what}, and MimiGen Stack D0 and Coffee D0 \cite{mandlekar23mimicgen} environments to a sparse reward setting, where the agent receives a reward of 0 until reaching a goal condition, which has a reward of 1. The environments have a fixed episode length: 100 in Pick cube and 400 in RoboMimic Lift, Can, and Square, and 800 in MimicGen Stack D0 and Coffee D0. We use a pair of shoulder and wrist RGB cameras as inputs to the RL agent; the reward predictors only use the shoulder camera. The action space in all environments consists of the delta motions of the end-effector (EE) at approximately 5 Hz as well as a gripper open/closed fraction command. All environments are instantiated with a Franka robot arm. In Pick cube, the EE rotation is fixed and in Lift, the EE rotation is unconstrained. We simplify the RoboMimic Can and Square environments by fixing the EE rotation in Can and limiting EE rotation to yaw only in Square. We use the provided demonstrations in Pick cube, Stack D0 and Coffee D0 and collect a new dataset of teleoperated demonstrations in Lift, Can, and Square using a 3Dconnexion SpaceMouse to teleoperate the robot. We also collect demonstrations in Lift and Square with a Kuka robot instead of a Franka.

\subsection{Real Environment Details}

\begin{figure}[t]
    \centering
    \includegraphics[width=1\linewidth]{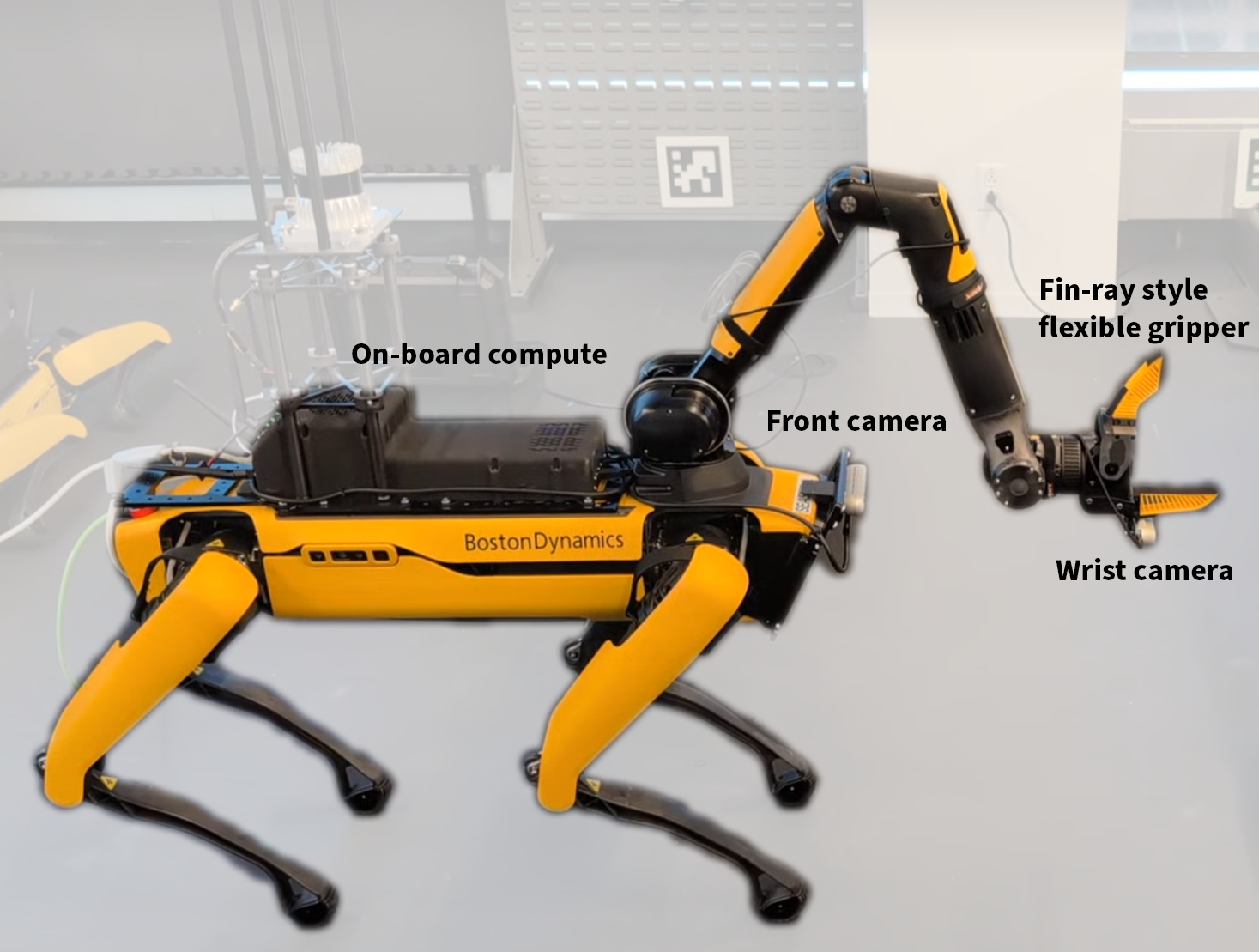}
    \caption{Our Boston Dynamics Spot setup used in the drawer opening experiment.}
    \label{fig:spot_setup}
\end{figure}

\begin{figure}[t]
    \centering
    \includegraphics[width=1\linewidth]{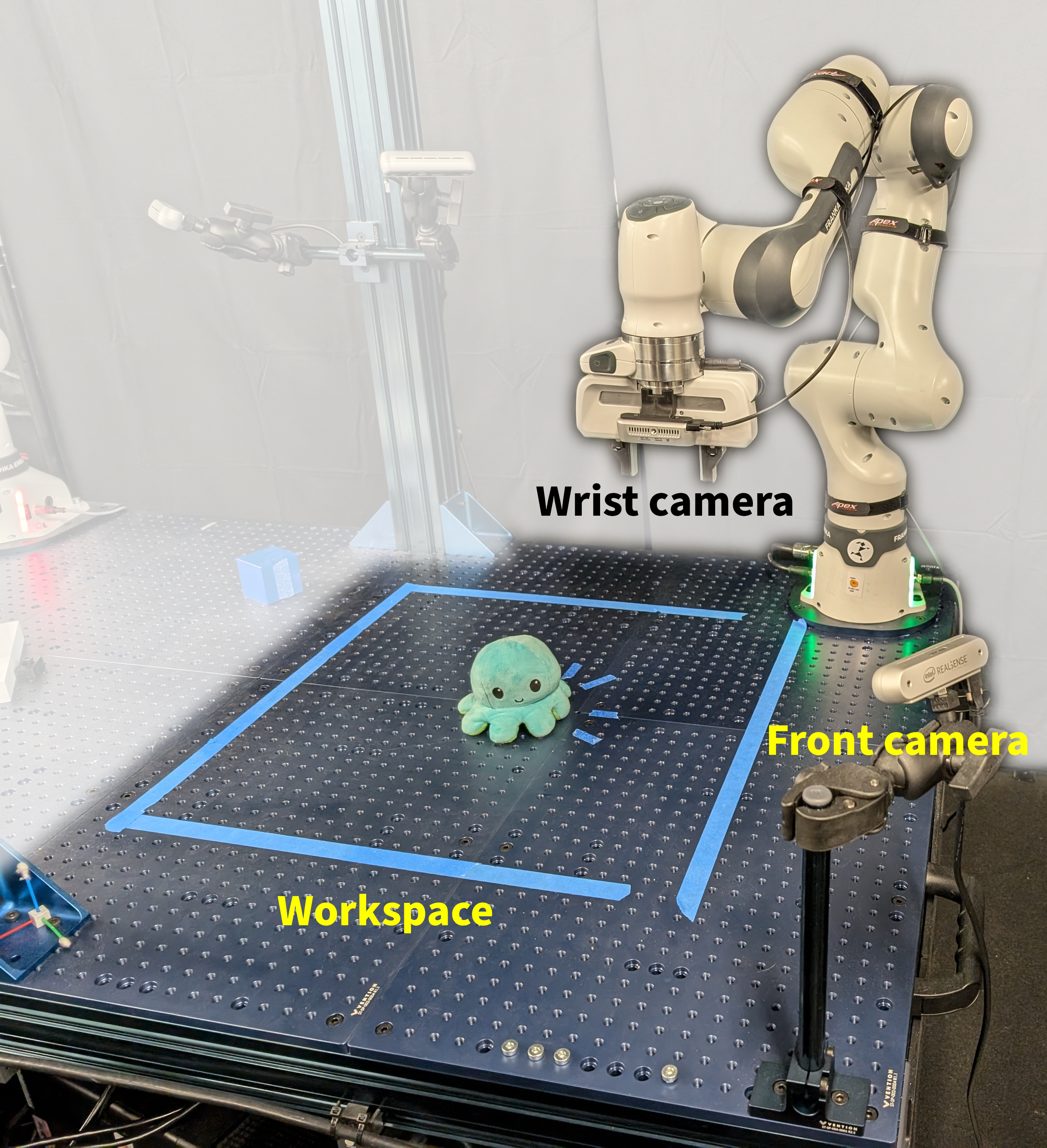}
    \caption{Our Franka tabletop setup used in the plushie picking and kettle opening experiments.}
    \label{fig:franka_setup}
\end{figure}

We perform real-world experiments with a table-top Franka arm (Figure \ref{fig:franka_setup}) and a Boston Dynamics Spot with a mounted arm (Figure \ref{fig:spot_setup}). The Franka task is to pick up a plushie and open a kettle by pushing a button. The Spot task is to open a drawer. We use a front-facing and a wrist-mounted camera in the Franka setup and a fixed EE rotation. The Spot setup involves two cameras mounted on the front of the Spot and the wrist of the Spot arm. In both cases, the front camera is a RealSense D455 and the wrist camera D435. The Spot uses a Fin-ray style soft gripper; the EE rotation is fixed to face the drawer. The Spot base moves in a semicircle around the drawer after each episode. The Franka and Spot tasks have a fixed episode length of 100 and 150 respectively. The Spot experiment has a scripted policy for opening a drawer to collect demonstrations and for closing the drawer for environment resets. The Spot automatically sits down and rolls over when its battery is at 10\% charge, after about an hour of running. We then swap the battery and resume the experiment. We use a SpaceMouse teleoperation setup to collect demonstrations with a Franka and reset its environment manually.

We also implement a goal classifier for each task, which provides extrinsic rewards to the reinforcement learning agent. For Spot drawer, we use an external camera to observe the drawer from the side. Then, we use Grounding DINO \cite{liu23grounding} to detect the drawer and Segment Anything \cite{kirillov23segment} to predict its segmentation mask. We assign a reward of 1 if the mask has increased by a certain size throughout the episode, which happens only when the drawer is open. For Franka plushie, we similarly detect and segment a plushie. Then, we threshold the height of the segmentation mask to predict success. For Franka kettle, we ask GPT-4o vision to answer a yes/no question for each frame with the following prompt: "Is the kettle open? Think it step by step but limit your answer to ONLY a word 'yes' or 'no'. There might be a robot gripper inside the view but you should focus on the kettle. If you are not sure, answer 'no.', no false positive is allowed. Only yes or no should appear in the answer. I will tip you \$500 for a correct answer."

\subsection{Implementation Details}

We re-implement LIV \cite{ma23liv} and a version of VIP \cite{ma23vip} that uses improvements from LIV (most importantly, cosine distance instead of $L_2$ distance on embeddings). We initialize LIV with a ResNet-50 CLIP model and further pre-train it on the EPIC-KITCHENS dataset \cite{damen18scaling}. We then use the VIP loss (LIV without language) to fine-tune the model on each task. All methods use the LIV backbone pre-trained on EPIC-KITCHENS and fine-tuned during online learning. We further run Rank2Reward with the LIV and R3M backbone, both frozen and unfrozen, and report the best result. We use a SERL implementation of DrQ and RLPD \cite{yarats21image} RL agents for all experiments. The agents are trained with a standard replay buffer and the Adam optimizer. We use the default hyper-parameters of DrQ and RLPD from SERL.

We train GCR and all baselines with a batch size of 32, a learning rate of $10^{-6}$, a weight decay of $10^{-3}$, and a negative buffer size of 20k steps. We use the GCR loss with $\omega_1 = \omega_2 = 1$. We further multiply both the intrinsic and extrinsic rewards by 10, as this appears to help the RL agent learn faster from the first few times it experiences a success. For simulated experiments, we use two NVIDIA V100 GPUs for each random seed. The first GPU runs the RL actor, RL learner and intrinsic reward predictor. The second GPU runs the intrinsic reward learner. We use two NVIDIA A6000 GPUs in our real-world experiments. In these experiments, we run the RL actor and RL learner on the first GPU and the rest -- extrinsic reward predictor, intrinsic reward predictor and intrinsic reward learner -- on the second GPU.

We list the hyper-parameters of GCR, baselines and the deep RL agent in Table \ref{tab:hyperparameters}.

\begin{table}[]
    \centering
    \caption{GCR and baseline hyper-parameters used in our simulation experiments.}
    \begin{tabular}{cc}
        \toprule
        Parameter & Setting \\
        \midrule
        \multicolumn{2}{c}{GCR} \\
        \midrule
        Batch size & 32 \\
        Learning rate & $10^{-6}$ \\
        Weight decay & $10^{-3}$ \\
        Negative buffer size & 20000 \\
        Negative alpha & 0.95 \\
        $\omega_1$ & 1 \\
        $\omega_2$ & 1 \\
        \midrule
        \multicolumn{2}{c}{VIP / LIV} \\
        \midrule
        Number of negative examples & 3 \\
        Alpha & 0.95 \\
        Metric & cosine \\
        Learning rate & $10^{-6}$ \\
        Weight decay & $10^{-3}$ \\
        Vision weight & 1 \\
        CLIP weight & 0 \\
        Language weight & 0 \\
        L1 regularization weight & 0 \\
        L2 regularization weight & 0 \\
        Discount & 0.98 \\
        \midrule
        \multicolumn{2}{c}{Rank2Reward (+ GAIL, VICE, AIRL)} \\
        \midrule
        Batch size & 256 (R3M Frozen) \\
         & 128 (LIV Frozen) \\
         & 32 (R3M Unfrozen) \\
         & 16 (LIV Unfrozen) \\
        Learning rate & $10^{-4}$ \\
        Use mixup loss & True \\
        Use state noise & True \\
        \midrule
        \multicolumn{2}{c}{DrQ / RLPD} \\
        \midrule
        Buffer size & 200000 \\
        Batch size & 256 \\
        UTD ratio & 4 \\
        Random start steps & 1000 \\
        Critic ensemble size & 10 \\
        Critic subsample size & 2 \\
        Discount & 0.96 \\
        Initial temperature & $10^{-2}$ \\
        Learning rate & $3 * 10^{-4}$ \\
        Soft target update rate & $5 * 10^{-3}$ \\
        Actor / learner steps & $\sim$ 0.75 \\
        \bottomrule
    \end{tabular}
    \label{tab:hyperparameters}
\end{table}

\end{document}